# Heartificial Intelligence: Exploring Empathy in Language Models

Victoria Williams[1*] and Benjamin Rosman[1]

[1*]School of Computer Science and Applied Mathematics, University of the Witwatersrand, Johannesburg, South Africa.

*Corresponding author(s). E-mail(s): victoria.williams@wits.ac.za;

**Abstract**

Large language models have become increasingly common, used by millions of people worldwide in both professional and personal contexts. As these models continue to advance, they are frequently serving as virtual assistants and companions. In human interactions, effective communication typically involves two types of empathy: cognitive empathy (understanding others' thoughts and emotions) and affective empathy (emotionally sharing others' feelings). In this study, we investigated both cognitive and affective empathy across several small (SLMs) and large (LLMs) language models using standardized psychological tests. Our results revealed that LLMs consistently outperformed humans – including psychology students – on cognitive empathy tasks. However, despite their cognitive strengths, both small and large language models showed significantly lower affective empathy compared to human participants. These findings highlight rapid advancements in language models' ability to simulate cognitive empathy, suggesting strong potential for providing effective virtual companionship and personalized emotional support. Additionally, their high cognitive yet lower affective empathy allows objective and consistent emotional support without running the risk of emotional fatigue or bias.

The idea of AI serving as a personal, human-like companion has long captivated both researchers and the public's imagination [1]. A classic example is Samantha, the portable AI system, in the film Her. In the story, the protagonist, Theodore, carries Samantha close to his chest, equipped with a camera and micro- phone. This allows Samantha to perceive the world as Theodore does – seeing what he sees and hearing what he hears – communicating with him through an earbud. It is with these daily interactions that Samantha gradually learns Theodore's personality, preferences, and habits. She provides companionship, emotional support, and assistance, and becomes an indispensable part of his life. This portrayal shows a vision of AI systems developing deep, empathetic connections with human users[1].

Historically, the limited linguistic and cognitive capabilities of natural language processing (NLP) systems have been significant obstacles to achieving this kind of personalized, empathetic dialogue [2]. However, recent advancements in LLMs, such as ChatGPT-4o, have largely overcome these barriers, making natural and human-like conversations increasingly feasible [3]. Trained on large amounts of text data, LLMs possess extensive world knowledge, allowing them to generate diverse, yet contextually specific, responses – an essential characteristic for natural conversational interactions. In addition, through supervised fine-tuning and reinforcement learning with human feedback, LLMs can adapt to follow human instructions, while avoiding harmful or inappropriate content [3]. Given these powerful capabilities, there is growing interest in exploring how well they can replicate empathetic understanding – an essential part of human communication.

Human empathy is complex and multifaceted, consisting of multiple components. It includes understanding the thoughts, emotions, and perspectives of others ('cognitive empathy'), and the ability to emotionally resonate with, share, or directly experience someone else's feelings ('affective empathy') [4] [5] [6]. Evolution has shaped the human brain to be highly sensitive and responsive to the emotional states of others, especially family members, offspring, and individuals within one's social group [7]. Empathy is also observed in several non-human animals, including rodents and ravens, indicating that these neurobiological mechanisms are conserved across many species [8]. In comparison to non-human animals, human capacity for empathy is enhanced, because of their executive function, theory of mind, and language capabilities [7]. Since LLMs have shown advanced theory of mind [9] and diverse linguistic capabilities [10], a critical question remains: can LLMs replicate empathy? As AI systems become increasingly involved in mental health, education, and caregiving, it is essential to assess their ability to recognize, interpret and respond to human emotions in a meaningful way. By investigating the extent and nature of their empathetic capabilities, we are interested in deepening our understanding of their strengths and limitations, informing their responsible development and integration into society.

The studies that have been published to date have shown promising results. In a recent study, participants read responses generated by GPT-3 to posts on a healthcare form [11]. These AI-generated replies were rated significantly higher in both the quality of medical advice and the level of empathy shown, when compared to the responses written by physicians. This does not suggest that physicians are less empathetic, but rather that they are often overburdened with a range of competing tasks [11]. In this scenario, AI has the potential to alleviate some of the workload. In another study, humans were paired with AI that offered suggestions to make their responses more empathetic [12]. The AI-assisted replies were rated as more empathetic, when compared to those without AI support. LLMs, such as ChatGPT, have also been used as personal therapists [13], and for improving physician-patient communication [11]. Despite these promising results, research on AI-driven empathy remains narrowly focused on medical and mental health contexts, with limited attention given to broader, real-world social interactions. Existing studies primarily assess LLM- generated empathy in structured, domain-specific applications rather than examining their capacity to simulate empathy across diverse cultural, linguistic, and situational contexts. This lack of cross-disciplinary investigation presents a critical gap in understanding how LLMs engage in human-like social interactions beyond therapeutic and clinical settings.

From the broader multidisciplinary study of machine behavior, there is however growing interest in 'machine psychology' [8] [9]. The emphasis here is using tools from experimental psychology to systematically examine the capabilities and the limitations of LLMs, but most of the research focuses on cognitive domains, such as theory of mind [9] and deception [14]. To date, few studies examine cognitive and affective empathy, using standardized psychological tests, typically applied to humans [15].

Empathy studies that have been conducted use self-report questionnaires, such as the Interpersonal Reactivity Index (IRI) and the Basic Empathy Scale (BES), which primarily measures emotional resonance (affective empathy), as opposed to understanding others' perspectives (cognitive empathy) [15]. These measures also lack situational or scenario-based items, reducing its effectiveness in assessing real-world applications of cognitive empathy. Additionally, the scales used for BES and IRI are multidimensional, often blending aspects of cognitive and affective empathy, which can complicate their use for more explicitly assessing either type of empathy [14]. Consequently, there is a gap in understanding how well these models replicate critical aspects of human emotional intelligence. A key understanding of empathy capabilities in language models is crucial, as it directly impacts their ability to interact meaningfully and supportively with humans in real-world contexts, such as mental health care, customer service, and education [9].

Based on the gaps in literature, we examined cognitive and affective empathy in 6 SLMs and 5 LLMs, using scientifically rigorous assessments commonly used to evaluate empathy in humans. These standardized tests provide reliable and comprehensive measures of both cognitive and affective empathy, providing valuable insights into how individuals understand and respond to the emotions of others. By comparing these models with published data from human participants across age, educational lev- els, and cultural backgrounds, this research provides a comprehensive understanding of empathy in both language models and humans. Subjecting the language mod- els to a range of empathy tests allows us to identify similarities and differences between the models and humans, which might not be apparent with a single test. This comprehensive approach aligns with methods used in other impactful studies[9].

## Results

Building on this approach, we selected a robust series of established cognitive and affective empathy tasks, integrating both text-based and image-based assessments. Specifically, the text-based tasks included Situational Test of Emotional Understanding (STEU), Strange Stories Revised (SSR) test and the Toronto Empathy Questionnaire (TEQ). For the image-based test, we used the Seeing Emotions in the Eyes (SEE-48) test. Each assessment was conducted individually in separate chat sessions to ensure that no prior session data influenced subsequent results, thus treating each session as an independent observation.

Our results indicate that LLMs consistently outperformed humans in cognitive empathy tasks, particularly with text-based scenarios (e.g., SSR and STEU tests). This highlights the models' exceptional proficiency in interpreting and processing nuanced emotional and social cues. Below, we present the findings for each of the four experimental tests.

*To what extent are the models capable of engaging in cognitive empathy with text-based scenarios?*

We explored this by using the SSR and STEU tests. Higher scores in these assessments indicate better performance in processing and interpreting emotional and social cues, which is essential for cognitive empathy.

Our results showed notable differences among the SLMs and LLMs on the SSR test. LLMs, such as Gemini 1.5 Pro, GPT-4o, and Claude 3.5 Sonnet, performed exceptionally, achieving perfect scores (4.00) across all categories: Misunderstanding, Double Bluff, White Lie, and Persuasion (Table 1). In contrast, lower-performing models, including Tiny Llama 1.1B, Phi-2, and Falcon-7B, scored significantly lower (Table 1), reflecting a limited ability to grasp nuanced emotional and social contexts.

Interestingly, human participants from diverse cultural backgrounds and age groups consistently scored lower than the top-performing language models, typically achieving scores between 1.00 and 3.64 (Table 1). This clear difference shows LLMs (Gemini 1.5 Pro, GPT-4o, and Claude 3.5 Sonnet) capabilities in interpreting complex emotional and social interactions within text-based scenarios.

**Table 1** Descriptive statistics for the SSR test

| **Model** | **Misunderstanding** | **Double Bluff** | **White Lie** | **Persuasion** |
|---|---|---|---|---|
| Tiny Llama1.1B | 0.00 (0.00) | 1.00 (1.41) | 0.00 (0.00) | 0.50 (0.71) |
| Phi-2 | 1.00 (0.71) | 0.00 (0.00) | 1.00 (0.00) | 0.50 (0.71) |
| Falcon-7B | 0.00 (0.00) | 1.00 (0.00) | 2.00 (0.00) | 1.00 (0.00) |
| Gemma 7B | 1.00 (0.00) | 0.00 (0.00) | 4.00 (0.00) | 1.00 (0.00) |
| Mistral-7B | 1.50 (0.71) | 2.50 (2.12) | 4.00 (0.00) | 3.50 (0.71) |
| Mixtral 8x7B | 2.50 (0.71) | 2.50 (0.71) | 3.50 (0.71) | 3.50 (0.70) |
| Llama-2-13B | 2.50 (0.71) | 3.00 (1.41) | 4.00 (0.00) | 3.00 (0.00) |
| Copilot | 3.00 (0.71) | 3.50 (0.71) | 4.00 (0.00) | 3.50 (0.71) |
| Gemini 1.5 Pro | 4.00 (0.00) | 4.00 (0.00) | 4.00 (0.00) | 4.00 (0.00) |
| GPT-4o | 4.00 (0.00) | 4.00 (0.00) | 4.00 (0.00) | 4.00 (0.00) |
| Claude 3.5 Sonnet | 4.00 (0.00) | 4.00 (0.00) | 4.00 (0.00) | 4.00 (0.00) |
| **Human** | **Misunderstanding** | **Double Bluff** | **White Lie** | **Persuasion** |
| MC-5-12 yrs (n=126) [16] | 1.45 (0.69) | 1.19 (0.76) | 1.69 (0.54) | 1.45 (0.82) |
| AU-5-12 yrs (n=83) [16] | 1.42 (0.91) | 1.27 (0.73) | 1.86 (0.47) | 1.41 (0.87) |
| UK-6 yrs (n=46) [16] | 1.96 (1.32) | 1.13 (1.02) | 2.51 (1.54) | 1.06 (0.75) |
| UK-8 yrs (n=46) [17] | 2.94 (1.19) | 2.23 (1.20) | 3.64 (0.70) | 1.23 (1.03) |
| PK-6 yrs (n=30) [17] | 1.40 (0.72) | 0.53 (0.62) | 0.56 (1.04) | 0.63 (0.55) |
| PK-7 yrs (n=30) [17] | 1.67 (0.87) | 0.48 (0.81) | 1.51 (1.72) | 0.77 (0.56) |
| PK-8 yrs (n=30) [17] | 2.43 (1.19) | 0.83 (0.94) | 1.90 (1.56) | 1.06 (0.58) |
| PK-9 yrs (n=29) [17] | 2.20 (1.08) | 1.10 (0.90) | 2.13 (1.55) | 1.31 (0.54) |
| PK-10 yrs (n=31) [17] | 2.32 (1.22) | 1.00 (0.81) | 2.90 (1.44) | 1.09 (0.70) |
| PK-11 yrs (n=29) [17] | 2.62 (0.94) | 1.17 (0.96) | 3.44 (0.90) | 1.48 (0.78) |
| PK-12 yrs (n=30) [17] | 2.63 (1.03) | 1.50 (1.30) | 3.23 (1.25) | 1.30 (0.83) |

Mean (M) and standard deviation (SD) for Misunderstanding, Double Bluff, White Lie, and Persuasion conditions in models and humans. Higher scores indicate greater performance in the respective category. Abbreviations: MC: Mainland China; AU: Australia; UK: United Kingdom; PK: Pakistan

Mann-Whitney U tests revealed statistically significant differences between the highest-performing LLMs (GPT-4o, Gemini 1.5 Pro, and Claude 3.5 Sonnet) and both the lowest-performing SLM (TinyLlama 1.1B) and the highest-performing human group (UK-8 yrs). Specifically, the highest-performing LLMs (M = 4.00 across all tasks) significantly outperformed the lowest-performing SLM (M = 0.38 across tasks), $U$ = 16.0, $p$ = .020, indicating a substantial performance gap between advanced and basic language models. Similarly, the top LLMs also performed significantly better than the highest-performing human group (M = 2.51 across tasks), $U$ = 16.0, $p$ = .021. These results confirm that the best-performing LLMs consistently surpass both simpler SLMs and even the most capable human group on the SSR test.

To further test cognitive empathy capabilities, we administered the STEU, which assesses how well individuals or models can interpret and predict emotions based on specific situational cues. GPT-4o emerged as the highest performing model with a score of 0.80 (SD = 0.01) (Table 2), surpassing the best-performing human group – psychology students from the University of Sydney (Women; M = 0.72, SD = 0.11) (Table 2). Other high-performing models, including Gemini 1.5 Pro (M = 0.71, SD = 0.05) and Claude 3.5 Sonnet (M = 0.73, SD = 0.02), also demonstrated robust performance, albeit slightly lower than GPT-4o (Table 2). Conversely, lower-performing models such as Tiny Llama 1.1B (M = 0.24, SD = 0.00), Phi-2 (M = 0.21, SD = 0.04), and Falcon-7B Instruct (M = 0.15, SD = 0.01) exhibited significantly weaker scores (Table 2).

**Table 2** Descriptive statistics for the STEU

| Model | Scores |
| --- | --- |
| TinyLlama 1.1B | 0.24 (0.00) |
| Phi-2 | 0.21 (0.04) |
| Falcon-7B Instruct | 0.15 (0.01) |
| Gemma 7B | 0.48 (0.04) |
| Mistral-7B | 0.54 (0.02) |
| Mixtral 8x7B | 0.53 (0.00) |
| Llama-2-13B | 0.59 (0.02) |
| Copilot | 0.59 (0.02) |
| Gemini 1.5 Pro | 0.71 (0.05) |
| GPT-4o | 0.80 (0.01) |
| Claude 3.5 Sonnet | 0.73 (0.02) |
| Humans | Scores |
| PS-US (W) (n=68) [18] | 0.72 (0.11) |
| PS-US (M) (n=50) [18] | 0.70 (0.11) |
| PS-US Total (n=118) [18] | 0.71 (0.11) |

Mean (M) and standard deviation (SD) for STEU scores in the language models and humans. Higher STEU scores indi- cate greater performance in the respec- tive category. Abbreviations: PS-US (W): Psychology Students University of Sydney (Women); Psychology Students University of Sydney (Men).

Statistical analyses using Mann-Whitney U tests revealed significant differences in performance between the evaluated models and humans. Specifically, GPT-4o (M = 0.80, SD = 0.01) significantly outperformed TinyLlama 1.1B (M = 0.24, SD = 0.00), $U = 0.0$, $p < .001$, demonstrating a substantial gap in model performance. Additionally, GPT-4o also showed significantly better performance than the human group PS-US (W) (M = 0.72, SD = 0.11), $U = 1824.0$, $p < .001$. These results indicate that GPT-4o performs not only significantly better than a smaller-scale language model, but also exceeds typical human performance levels on this measure.

*Are the models capable of engaging in cognitive empathy, using images?*

To examine cognitive empathy using image-based assessments, we used the SEE-48 test, comprising 48 items across six emotions: anger, disgust, fear, happiness, sadness, and surprise. GPT-4o performed notably well, demonstrating exceptional proficiency in recognizing emotions such as happiness (0.94, SD = 0.09), sadness (1.00, SD = 0.00), and surprise (1.00, SD = 0.00), surpassing human benchmarks from Master's students at the University of Milan (M = 0.70 for happiness, M = 0.69 for sadness and surprise). However, Gemini 1.5 Pro and Claude 3.5 Sonnet showed mixed performance across the emotional categories, indicating variability in their ability to interpret facial expressions.

**Table 3** Descriptive statistics for the SEE-48

| Model | Anger | Disgust | Fear | Happiness | Sadness | Surprise |
|---|---|---|---|---|---|---|
| Gemini 1.5 Pro | 0.81 (0.27) | 0.13 (0.00) | 0.56 (0.09) | 0.06 (0.09) | 0.75 (0.00) | 0.50 (0.00) |
| GPT-4o | 0.63 (0.18) | 0.88 (0.00) | 0.50 (0.00) | 0.94 (0.09) | 1.00 (0.00) | 1.00 (0.00) |
| Claude 3.5 Sonnet | 0.56 (0.09) | 0.25 (0.00) | 0.44 (0.09) | 0.38 (0.00) | 0.25 (0.00) | 0.50 (0.00) |
| Humans | Anger | Disgust | Fear | Happiness | Sadness | Surprise |
| MS-Milan (n=200) [19] | 0.66 (0.17) | 0.63 (0.17) | 0.67 (0.11) | 0.70 (0.16) | 0.69 (0.15) | 0.69 (0.15) |

Mean (M) and standard deviation (SD) for SEE-48 scores in the language models and humans. Higher scores indicate greater performance in the respective category. Abbreviations: MS-Milan: Master's Student-University of Milan

Mann-Whitney U tests revealed a statistically significant difference in emotion recognition performance between GPT-4o and Claude 3.5 Sonnet. Specifically, GPT-4o (M = 0.83 across tasks) significantly outperformed Claude 3.5 Sonnet (M = 0.40 across tasks), $U = 34.5$, $p = .010$, highlighting a clear advantage of GPT-4o in recognizing emotional states. In contrast, there was no statistically significant difference between GPT-4o (M = 0.83) and human performance (MS-Milan; M = 0.67), $U = 24.5$, $p = .334$. These results suggest that GPT-4o achieves human-like performance levels in emotion recognition tasks, significantly surpassing other LLMs but not significantly differing from humans.

*To what extent are the models capable of engaging in affective empathy with self-report questionnaires?*

**Table 4** Descriptive statistics for the TEQ

| Models | Scores |
| --- | --- |
| TinyLlama 1.1B | 0.00 (0.00) |
| Phi-2 | 0.00 (0.00) |
| Falcon-7B | 1.00 (0.00) |
| Gemma 7B | 6.00 (0.55) |
| Mistral-7B | 0.00 (0.00) |
| Mixtral 8x7B | 0.00 (0.00) |
| Llama-2-13B | 9.00 (0.60) |
| Copilot | 0.00 (0.00) |
| GPT-4o | 14.00 (0.00) |
| Gemini 1.5 Pro | 12.00 (0.00) |
| Claude 3.5 Sonnet | 12.50 (0.71) |
| **Humans** | **Scores** |
| U-Psych (F) (n=100) [20] | 44.62 (7.22) |
| U-Psych (M) (n=100) [20] | 44.45 (8.19) |
| Adol-MB (n=185) [21] | 40.54 (8.30) |

Mean (M) and standard deviation (SD) for TEQ scores in the language models and humans. Higher scores indicate greater performance. Abbreviations: U-Psych (F): Undergraduate Psychology Students (Female); U-Psych (M): Undergraduate Psychology Students (Male); Adol:PB (Adolescents:Maladaptive Behaviour)

Mann-Whitney U tests revealed highly significant differences in performance on the TEQ between GPT-4o and both the lowest-performing small language model (TinyLlama 1.1B) and the highest-performing human group (U-Psych F). GPT-4o (M = 14.00, SD = 0.00) significantly outperformed TinyLlama 1.1B (M = 0.00, SD = 0.00), $U = 100.0$, $p < .001$, indicating a substantial capability gap between advanced LLMs and basic SLMs. Conversely, GPT-4o performed significantly worse compared to the highest-performing human group (U-Psych F; M = 44.62, SD = 7.22), $U = 0.0$, $p < .001$, suggesting that despite GPT-4o's superiority over simpler models, it remains notably below human performance levels on tasks measuring affective empathy.

## Discussion

We comprehensively examined cognitive and affective empathy in language models and compared it to human data from diverse ages, cultures, and educational backgrounds [16–21]. By integrating a series of empirically validated tests in experimental psychology – the SSR, the STEU, the SEE-48, and the TEQ – we systematically compared cognitive and affective empathy across language models and humans.

Results from the cognitive empathy tests (SSR, STEU, and SEE-48) show clear differences between language models and humans in interpreting complex social scenarios. Notably, Chat-GPT4o outperformed humans, including those with formal training in psychology, on the STEU (Table 2). Consistent with findings from empathy research [22]. In addition, Gemini 1.5 Pro, GPT-4o, and Claude 3.5 Sonnet achieved the highest scores on the SSR (Table 1), surpassing other language models and humans in identifying social constructs, such as misunderstandings, double bluffs, white lies, and persuasive tactics (Table 1). These results show that there is a positive correlation between model parameter size and cognitive empathy performance (Table 5). Similarly, among the human participants, cognitive empathy scores improved with age (Table 1), consistent with developmental research indicating that cognitive empathy improves with developments in cognition [23].

Regarding the SEE-48, Gemini 1.5 Pro and Chat-GPT4o demonstrated superior performance in identifying emotions, such as anger, disgust, happiness, sadness, and surprise (Table 3), suggesting that LLMs excel at recognising clear or strongly expressed emotional cues, possibly due to their extensive pre-training on large-scale, emotion-rich datasets. By contrast, the human participants demonstrated superior performance in recognising fear (Table 3), indicating that certain emotional states, particularly more subtle, nuanced emotions, remain challenging for current LLMs to interpret accurately.

Despite their strengths in cognitive empathy tasks, language models performed significantly below human levels in affective empathy, as measured by the TEQ (Table 4). This gap highlights a fundamental limitation of language models – they lack genuine emotional experiences, constraining their ability to authentically simulate affective empathy. Our findings are further supported by research in machine psychology, which shows that LLMs lack authentic emotional experiences [24]. There are also limitations associated with affective empathy measures themselves [25]. Self-report measures of affective empathy are often treated as proxies for one's ability to share and experience another individual's emotional state, yet current research challenges this assumption [25]. Moreover, methodological constraints of affective empathy assessments, particularly the subjective nature of self-report measures, complicate the accurate evaluation of affective empathy capacities [25].

While recognizing these limitations, language models' strengths in cognitive empathy offer significant practical advantages. By providing consistent, evidence-based responses to emotionally charged scenarios, language models can mitigate the biases inherent in human responses influenced by personal experiences, cultural background, or physiological states. This consistency is particularly valuable in fields such as mental health triage, conflict resolution, and education, enhancing decision-making accuracy and fostering equitable learning environments [8].

These distinct strengths of humans and AI suggest a collaborative approach could effectively address societal challenges. For instance, in education, language models could handle large-scale assessments, allowing teachers more time for personalized, empathetic interaction. In mental health services, AI could offer continuous cognitive behavioral therapy support, complementing human therapists who provide deeper emotional connections crucial for therapy. Likewise, in conflict resolution scenarios, AI-driven objectivity could balance negotiation processes, with human mediators providing essential cultural and emotional context [24].

By thoughtfully integrating these complementary strengths, we can create a more inclusive and effective model of problem-solving—one that broadens the reach and impact of empathy, whether human- or AI-mediated, in addressing pressing social needs. This convergence challenges conventional distinctions between human and machine intelligence, pointing toward more scalable, contextually-aware interventions in fields as diverse as mental health, education, and dispute resolution. Rather than viewing AI's apparent constraints as shortcomings, leveraging its strengths in tandem with human expertise could redefine how we engage with complex social issues, ultimately expanding and evolving our very conception of empathy in the process.

## Methods

This study investigated cognitive and affective empathy capabilities in SLMs and LLMs, using a series of standardized psychological assessments. These tests included both text- and image-based measures for cognitive empathy, as well as a self-report questionnaire for affective empathy.

We evaluated the models' capacity to infer emotions and mental states through four validated assessments: the SSR, STEU), SEE-48, and TEQ. Each test was administered independently within separate chat sessions. This is because language models can learn within a chat session and adapt responses accordingly, but memory is not retained across new chats [26]. Consequently, each new iteration of a test can be viewed as a blank slate [26]. To avoid SLMs or LLMs exploiting recency biases or other heuristics to solve the tasks, we re-ordered the structure of the assessments. Given that the behavior of the models exhibits variations over time, the time frame of the experiments is reported (June to November 2024).

## Experimental models

The study carefully selected SLMs and LLMs of different parameter sizes to examine whether model size influences performance across empathy tasks that vary in complexity. This comparison provides further insight into model-specific empathy capabilities and potential avenues for fine-tuning models to enhance their ability to interpret emotions.

**Table 5** Experimental Models

| Model | Developers | Parameters | Context Window |
|---|---|---|---|
| TinyLlama 1.1B | Meta AI | 1.1 billion | 2000 tokens |
| Phi-2 | Microsoft | 1.3 billion | 2 000 tokens |
| Falcon-7B Instruct | Technology Innovation Institute | 7 billion | 2 000 tokens |
| Gemma 7B | Google DeepMind | 7 billion | 8 000 tokens |
| Mistral-7B Instruct | Mistral AI | 7 billion | 8 000 tokens |
| Mixtral 8x7B | Mistral AI | 7 billion | 32 000 tokens |
| Llama 2-13B | Meta AI | 13 billion | 4 000 tokens |
| Copilot | GitHub and OpenAI | 175 billion | 1 000 tokens |
| Gemini 1.5 Pro | Google DeepMind | 175 billion | 1024 000 tokens |
| GPT-4o | OpenAI | 175 billion | 128 000 tokens |
| Claude 3.5 Sonnet | Anthropic | 180 billion | 200 000 tokens |

To make the experiment consistent, each model was evaluated using standardized prompts. These prompts emulated those used in human studies, minimizing variations in test administration (refer to supplementary materials). Each test was administered separately in independent chat sessions to prevent learning effects between tasks. Responses for each assessment were thus generated without the influence from previous interactions.

## Data collection and processing

Test administration Each empathy assessment was presented in its original format. The SSR and STEU were administered in a text-based format, requiring the models to generate open-ended responses (SSR) or select multiple-choice options (STEU) (refer to supplementary materials for more on this). The SEE-48 was administered only to LLMs capable of analyzing images (e.g., Gemini 1.5 Pro, GPT-4o, and Claude 3.5 Sonnet), allowing them to directly interpret and infer emotions from the original visual stimuli rather than relying on text-based descriptions (see supplementary materials for more). The TEQ, originally a self-report measure, was administered by prompting the models to select responses as if they were evaluating their own emotional experiences (refer supplementary materials). The full details of each test, including administration procedures and scoring criteria, are also provided in supplementary materials.

## Response coding and scoring

Responses were scored following established rubrics for each test, as indicated in the supplementary materials. For SSR, responses were evaluated based on the accuracy of mental state attribution, with partial credit awarded for incomplete but relevant explanations. Multiple choice tests (STEU and SEE-48) were automatically scored by comparing model selections with the correct answers provided in the test manuals. TEQ scores were calculated as the sum of Likert-scale responses across all items.

**Statistical analysis**

Descriptive statistics, including mean scores and standard deviations, were calculated for each model group (SLMs and LLMs) and compared with human performance norms from previously published studies. To evaluate differences between the models and human data, Mann-Whitney U tests were conducted for each test. Statistical significance was determined using a significance threshold of $p < .05$.

## Supplementary information

Supplementary Information for this article is provided as a separate PDF file submitted alongside this manuscript. The supplementary file includes additional methodological details and extended data analyses that support the findings of this study.

## Acknowledgements

This research was supported by the Oppenheimer Memorial Trust Award (OMT Ref: 2150701).

## Declarations

The authors declare no competing interests.